\documentclass{article}

\usepackage[final]{corl_2026} % Uncomment for the camera-ready ``final'' version.

\usepackage{graphicx}
\usepackage{amssymb}
\usepackage{amsmath}
\usepackage{bm}
\usepackage{makecell, booktabs, multirow, tabularx}
\usepackage{diagbox}
\usepackage{enumitem}
\usepackage[table]{xcolor}
\usepackage{subcaption}

\usepackage{xcolor}

\title{
Spatially Conditioned Diffusion Policy: Learning Precise and Robust Manipulation with a Single RGB Camera}

\author{
    Seoyoon Kim$^{1,*}$ \quad
    Kanghyun Kim$^{1,*}$ \quad
    Dongwoo Ko$^{2}$ \quad
    Young Jin Heo$^{2}$ \quad
    Min Jun Kim$^{1}$\\[0.2em]
    $^{1}$Korea Advanced Institute of Science and Technology (KAIST) \\
    $^{2}$Neuromeka\\
    {\small $^{*}$Equal contribution.} \\
    \url{https://scdp-project.github.io/}\\
}

\begin{document}
\maketitle
%===============================================================================

\begin{figure}[h]
    \centering
    \vspace{-2mm}
    \includegraphics[width=\linewidth]{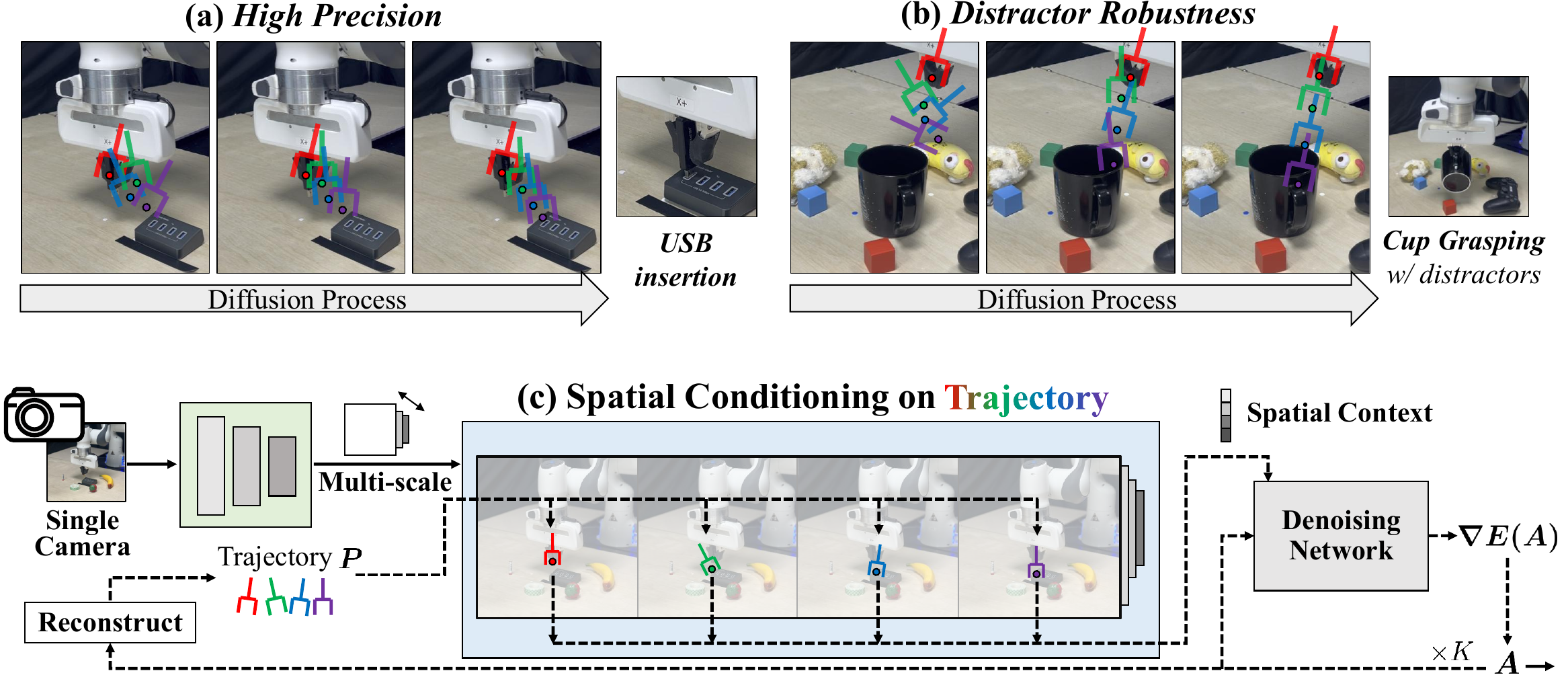}
    \caption{
    (a, b) In a single-camera setting, precise and robust manipulation requires the robot to capture fine-grained visual cues while identifying task-relevant regions within a single RGB image. (c) To address this challenge, Spatially Conditioned Diffusion Policy (SCDP) uses intermediate denoising trajectories as spatial attention anchors on multi-scale image feature maps, allowing the policy to extract task-relevant visual features while preserving fine visual details.}
    \vspace{-2mm}
    \label{fig:main_cover}
\end{figure}

\begin{abstract}
Recent visual imitation learning systems have widely adopted multi-camera setups with wrist-mounted cameras as the de facto standard. However, manipulation from a single global view remains challenging, as the policy should capture fine-grained interaction details and identify task-relevant regions without local wrist views. To address this challenge, we present Spatially Conditioned Diffusion Policy (SCDP), a diffusion-based visuomotor policy that achieves precise and robust manipulation in a single-camera setting. Our key idea is that end-effector trajectories can serve as visual attention anchors that reflect task-relevant regions. Building on this idea, SCDP consists of two key components: (i) a visual encoder that produces multi-scale feature maps to capture both broader context and fine-grained visual features, and (ii) a spatial conditioning module that samples point-wise features along intermediate end-effector trajectories in the diffusion loop. Extensive simulation experiments show that SCDP consistently outperforms strong single-view baselines and achieves performance comparable to multi-camera baselines. Real-world experiments further demonstrate precise manipulation and robustness to visual distractors, highlighting the potential of single-camera imitation learning. Code and videos are available \href{https://scdp-project.github.io/}{here}.
\end{abstract}


%===============================================================================

\section{Introduction}
\label{sec:intro}
Imitation learning has emerged as a practical paradigm for robotic manipulation. By learning directly from expert demonstrations, imitation learning provides a scalable way to acquire complex policies without requiring explicit reward design or hand-crafted control pipelines~\citep{argall2009survey, osa2018algorithmic}.

In particular, recent works on visual imitation learning~\citep{chi2023diffusionpolicy, zhao2023learning, chi2024universal, zhao2024aloha} have demonstrated strong performance across diverse manipulation tasks, including real-world problems that require contact-rich interaction and precise control. 
In these settings, multi-camera setups combining a global view with wrist-mounted cameras have become a standard configuration. 
Notably, wrist cameras play a crucial role in this setup.
Their close-range views enable more detailed observation of local regions, thereby reducing visual ambiguity and allowing the policy to focus on the most relevant aspects of the task.

However, achieving precise manipulation from a single global view remains challenging. With only a single global view, the policy should capture fine-grained visual details required for task success (e.g., USB insertion, Fig.~\ref{fig:main_cover}(a)), while also identifying task-relevant regions within unstructured scenes (e.g., cup grasping with distractors, Fig.~\ref{fig:main_cover}(b)). Existing visuomotor policy frameworks typically compress observations into global feature embeddings~\citep{chi2023diffusionpolicy, jang2022bc} or decompose them into visual tokens~\citep{zhao2023learning, zhao2024aloha}. However, they do not explicitly capture fine-grained visual details or select task-relevant regions, requiring the policy to infer visual relevance implicitly.

%Several recent methods improve policy performance in a single-camera setup by explicitly emphasizing task-relevant regions in the visual observation. 
This challenge connects to recent efforts to learn efficient representations by emphasizing task-relevant regions in visual observations.
% For instance, SKIL~\citep{wang2025skil} uses a segmentation mask from~\citep{kirillov2023segment} to de-emphasize task-irrelevant regions, while OTTER~\citep{huang2025otter} uses a VLM to select only task-relevant visual features and passes them to the policy. 
For instance, SKIL~\citep{wang2025skil} uses a segmentation mask~\citep{kirillov2023segment} to identify semantic keypoints, while OTTER~\citep{huang2025otter} uses a VLM to attend to language-relevant regions.
%HODOR~\citep{qian2024task} leverages LLMs such as GPT-4~\citep{achiam2023gpt} to identify task-relevant entities, which are then segmented with Grounded SAM~\citep{ren2024grounded} to extract object-centric embeddings.
While effective, their reliance on large pretrained models complicates the pipeline and may be suboptimal for fine-grained control. Another line of work, such as DP3~\citep{ze20243d} and RISE~\citep{wang2024rise}, has incorporated 3D point-cloud representations to enable richer geometry information in a single-camera setup. However, these representations are also summarized into a global feature embedding, rather than explicitly focusing on task-relevant regions or capturing fine-grained geometric details. 

In this paper, we introduce \textbf{Spatially Conditioned Diffusion Policy (SCDP)}, which enables precise and robust manipulation using a single RGB camera. Our key insight is that future end-effector trajectories projected onto the image can provide useful attention anchors for identifying task-relevant regions. During the diffusion process, SCDP leverages intermediate action trajectories as evolving estimates of future motion and extracts point-wise features at those anchors (Fig.~\ref{fig:main_cover}(c)). Moreover, SCDP utilizes multi-scale image encoding to capture both fine local details and broader context.

% In summary, the main contributions of this paper are as follows: 1) We present \textbf{SCDP}, which enables precise and robust manipulation from a single global view, without relying on wrist-mounted cameras. 2) We propose a novel mechanism for scene conditioning that uses evolving action trajectories to sample point-wise visual features at task-relevant locations. 3) We integrate multi-scale image encodings to capture both coarse scene context and fine-grained visual details required for precise manipulation.
In summary, the main contributions of this paper are as follows: 

\begin{itemize}
    \item We present \textbf{SCDP}, which enables precise and robust manipulation from a single global view, without relying on wrist-mounted cameras.
    \item We propose a novel mechanism for scene conditioning that uses evolving action trajectories to sample point-wise visual features at task-relevant locations.
    \item We integrate multi-scale image encodings to capture both coarse scene context and fine-grained visual details required for precise interaction.
\end{itemize}

We empirically validate SCDP through extensive simulation experiments and real-world evaluations. With only 20 demonstrations, SCDP achieves over 80\% success across all Meta-World~\citep{yu2020meta} difficulty groups using a single global view. In real-world experiments, SCDP maintains high accuracy on high-precision tasks and robustness in distractor-rich settings. These findings demonstrate that a single global view can support effective visual representations for precise manipulation.

\section{Related Work}
\label{sec:related}

    %For example, Diffusion Policy~\citep{chi2023diffusionpolicy} and 3D Diffusion Policy~\citep{ze20243d} compress the whole scene into a single global scene vector, while ACT~\citep{zhao2023learning} and RISE~\citep{wang2024rise}represents observations as visual tokens processed by a transformer. 

\paragraph{Imitation Learning for Precise Manipulation.}
Many prior works address precise manipulation using multi-view observations, often including wrist-mounted cameras to provide task-relevant geometric information~\citep{chi2023diffusionpolicy, zhao2023learning, zhao2024aloha, fu2024mobile, lan2025bfa, saxena2024mrest}. 
Representative imitation learning frameworks such as Diffusion Policy~\citep{chi2023diffusionpolicy}, ACT~\citep{zhao2023learning}, and ALOHA Unleashed~\citep{zhao2024aloha} leverage multi-view visual inputs to model precise, contact-rich manipulation behaviors.
%In another line of work, RVT~\citep{goyal2023rvt} and RVT-2~\citep{goyal2024rvt} construct point-cloud representations from RGB-D observations and render them from multiple virtual viewpoints around the robot workspace for precise manipulation.
%More recent studies address high-precision manipulation, such as charger plugging and USB insertion, using multi-view observations together with human-in-the-loop reinforcement learning or RL-augmented teleoperation~\citep{tang2026towards, luo2025precise}.
Other works incorporate additional sensing modalities, such as force and tactile sensing, to leverage contact feedback for fine alignment and insertion~\citep{huang2025tactile,yu2026forcevla,wu2025tacdiffusion}. 
For example, ForceVLA~\citep{yu2026forcevla} performs USB insertion by combining world- and wrist-view images with external force sensing.
In contrast, SCDP aims to solve precision manipulation tasks, such as USB insertion, using only a single RGB image.

%In another line of work, RVT~\citep{goyal2023rvt} and RVT-2~\citep{goyal2024rvt} use RGB-D observations to reconstruct scene point clouds, which are then re-rendered from multiple virtual viewpoints around the robot workspace for precise manipulation.

% DP3~\citep{ze20243d} and RISE~\citep{wang2024rise} demonstrate the feasibility of single-view manipulation through efficient 3D point-cloud representations. 
% In contrast, SCDP enables precise manipulation using only a single RGB camera.

%Beyond vision-based approaches, some works incorporate additional sensing modalities to obtain complementary physical feedback. Force- and tactile-based approaches utilize contact feedback for fine alignment and insertion~\citep{huang2025tactile,yu2026forcevla,fu2023safe,wu2025tacdiffusion,kamijo2026tactile,tang2026towards}.

%Force- and tactile-based approaches utilize contact feedback for fine alignment and insertion~\citep{huang2025tactile,yu2026forcevla,fu2023safe,wu2025tacdiffusion,kamijo2026tactile, tang2026towards}.
%Other methods use explicit object pose information to model source-target relative poses~\citep{sun2025exploring}.
%Multi-view approaches improve access to task-relevant geometric information by leveraging multiple cameras~\citep{luo2025precise, saxena2024mrest, zhao2023learning, zhao2024aloha, lan2025bfa} or rendered multi-view representations~\citep{goyal2024rvt}.

% focus visual feature extraction on behaviorally relevant regions and 
% \paragraph{Imitation Learning with Efficient Representations.}
\paragraph{Task-Relevant Visual Conditioning.}
Recent works have explored conditioning downstream policies on additional task-relevant information to provide more informative cues for action prediction.
Existing approaches leverage large-scale pretrained models~\citep{nair2022r3m, ma2022vip, qian20253d, di2024dinobot}, extract task-relevant keypoints~\citep{wang2025skil, fang2025kalm}, build object-centric embeddings~\citep{qian2024task, shi2024composing, hsu2025spot, zhu2023viola, zhu2023learning}, or use vision-language models to identify task-relevant regions~\citep{huang2025otter, stone2023open}. While pretrained representations provide strong general-purpose features, they may encode the full scene, including task-irrelevant entities, without explicit spatial selection. In contrast, explicit task-relevant feature extraction methods can provide more spatially selective representations. 
However, they often rely on multi-stage perceptual pipelines, such as language-model prompting~\citep{achiam2023gpt}, segmentation~\citep{kirillov2023segment}, or point tracking modules~\citep{karaev2024cotracker}, where errors from early perception stages can cascade into downstream policy learning. 
SCDP instead leverages intermediate action trajectories within the diffusion loop as spatial queries to extract task-relevant features for subsequent denoising, without relying on external perception modules.

\paragraph{Multi-Scale Visual Representations.} 
Given that different levels of granularity are essential for diverse downstream tasks, prior works have explored hierarchical representations in both 2D domains~\citep{lin2017feature, cai2016unified, wang2021pyramid, xie2021segformer} and 3D point-cloud processing~\citep{qi2017pointnet++, wang2019dynamic, nie2022pyramid}. 
These approaches typically construct a pyramidal feature hierarchy to yield representations across multiple spatial scales, where lower-resolution features capture high-level semantic context and higher-resolution features preserve the fine-grained details. 
Inspired by the Feature Pyramid Network (FPN) \citep{lin2017feature}, SCDP adopts a multi-scale visual representation to capture information at diverse granularities from a single view.

\section{Method}
\label{sec:method}

\begin{figure}[t]
    \centering
    \includegraphics[width=\linewidth]{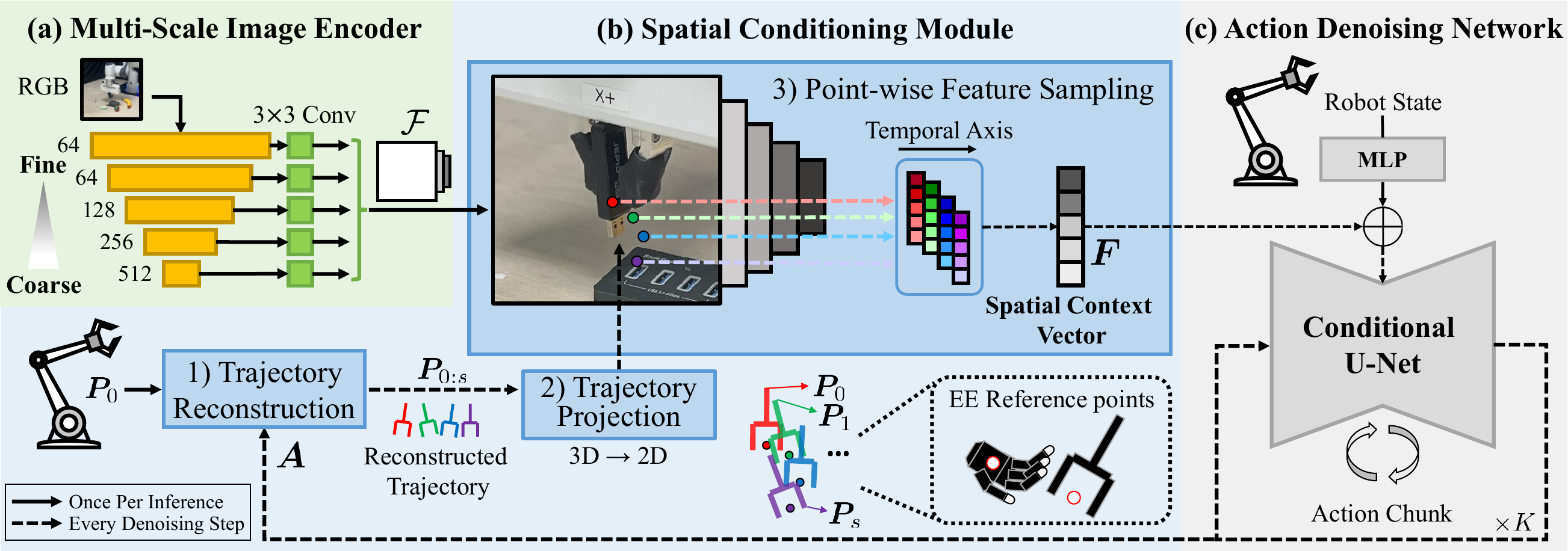}
    \caption{\textbf{SCDP Architecture.}
    (a) SCDP constructs hierarchical feature maps using a multi-scale image encoder.
    (b) Given the current robot state and intermediate action samples, SCDP reconstructs future trajectories and projects them onto multi-scale feature maps to extract localized visual features.
    (c) The features are aggregated into a spatial context vector, which conditions the diffusion U-Net together with the robot state to denoise action samples and predict the final action chunk.}
    \label{fig:SCDP_Framework}
    \vspace{-5mm}
\end{figure}

The overall architecture of the proposed SCDP is illustrated in Fig.~\ref{fig:SCDP_Framework}. SCDP primarily comprises three components: (i) a multi-scale image encoder that extracts coarse-to-fine feature maps from raw RGB images (Fig.~\ref{fig:SCDP_Framework}(a)), (ii) a spatial conditioning module that samples point-wise visual features using an intermediate action trajectory (Fig.~\ref{fig:SCDP_Framework}(b)), and (iii) an action denoising network that denoises the action trajectory conditioned on the extracted visual features and the current state (Fig.~\ref{fig:SCDP_Framework}(c)).

\subsection{Multi-Scale Image Encoder}
\label{subsec:multi_scale}

SCDP leverages a set of multi-scale feature maps to capture both global scene context and fine-grained visual details required for precise interaction. We use ResNet-18~\citep{he2016deep} as our backbone, denoting the stem output after max pooling as $C_1$ and the outputs of the four residual stages as $C_2, \dots, C_5$.
These feature maps have strides of 4, 4, 8, 16, and 32, trading spatial resolution for richer semantic features.
We then apply independent $3 \times 3$ convolutions to each $C_i$. 
These convolutions align the channel dimensions across different feature maps to $d$ and provide smoother local features for the subsequent denoising process.
This yields $\mathcal{F} = \{\mathcal{F}_1, \dots, \mathcal{F}_5\}$, where $\mathcal{F}_i \in \mathbb{R}^{d \times H_i \times W_i}$ denotes the feature map derived from the corresponding backbone feature $C_i$.
% As a result, we obtain a set of multi-scale feature maps $\mathcal{F} = \{\mathcal{F}_1, \dots, \mathcal{F}_5\}$, where each $\mathcal{F}_i \in \mathbb{R}^{d \times H_i \times W_i}$ denotes the feature map of the corresponding stage $C_i$.

\subsection{Spatial Conditioning Module}
\label{subsec:spatial_conditioning}
Our key insight is that future end-effector trajectories can provide visual attention anchors that reflect task-relevant regions. To leverage this insight, the module (i) reconstructs the end-effector point trajectory from an intermediate action trajectory $\bm{A}_t^k$ at denoising step $k$, (ii) projects the reconstructed trajectory onto the image plane, and (iii) samples point-wise features from the multi-scale feature maps $\mathcal{F}$. 
%Here, given current timestep $t$, the action trajectory $\bm{A}_t^k=\bm{a}^k_{t:t+s}$ denotes the sequence of actions. 
For simplicity, we omit the current timestep $t$ and diffusion step $k$ in this section.

\textbf{Trajectory Reconstruction.\,} Given a diffusion policy with a prediction horizon $h$, we define a reconstruction horizon $s$ ($s \leq h$) to represent the reconstructed future end-effector (EE) trajectory. 
We parameterize actions as a relative displacement from the current state.
Therefore, the trajectory estimate can be derived from the current position and the action sequence. 
We construct a point trajectory $\bm{P}_{0:s} = (\bm{P}_0, \bm{P}_1, \dots, \bm{P}_s)$, where $\bm{P}_i \in \mathbb{R}^3$ for $i \in \{1, \dots, s\}$ denotes the position of the EE reference point at timestep $i$, and $\bm{P}_0$ indicates the current EE position. 
We extract the translational displacements $\Delta \bm{P}_{1:s}$ from the current action sequence $\bm{A} = \bm{a}_{1:h}$ at denoising step $k$, where $\bm{a}_{i}$ denotes a predicted action element at timestep $i$.
The future positions are reconstructed using a cumulative formulation: $\bm{P}_{i} = \bm{P}_{i-1} + \Delta \bm{P}_i$. While $\bm{P}_0$ remains constant as determined by the current state, the subsequent trajectory points $\bm{P}_{1:s}$ are iteratively updated throughout the denoising process. This allows the model to dynamically adjust its spatial attention to task-relevant regions.
% Figure~\ref{fig:ablation}(a) illustrates the effect of spatial conditioning as $s$ increases.
    % This distinction allows the model to maintain a fixed grounding in the robot's present configuration while adaptively adjusting its spatial attention on the evolving future path.

% \textbf{Trajectory Projection onto the Image Plane\,} As the point trajectory $\bm{P}_t^k$ consists of 3D points in task space, we project its points onto the image plane to determine the locations of the attention anchors. For each $i \in \{0, \dots, s\}$, the point $\bm{p}_{t+i}^k \in \bm{P}_t^k$ is transformed into the camera coordinate frame via the extrinsic matrix $[\bm{R} | \bm{t}] \in SE(3)$, which represents a rigid-body transformation: $\bm{p}^{cam}_{t+i} = \bm{R} \,\bm{p}_{t+i}^k + \bm{t}$. The transformed 3D points are then projected into 2D pixel coordinates $(u_i, v_i)_{i=0}^{s}$ using the camera intrinsic matrix $\bm{K}$. 

% \subsubsection{Spatial Projection}
% \label{subsubsec: spatial_projection}
\textbf{Trajectory Projection.\,} Since the trajectory sequence $\bm{P}_{0:s}$ consists of 3D points in task space, we project them onto the image plane to determine the locations of the attention anchors. 
Each point $\bm{P}_i$ in the robot base frame is transformed into the camera coordinate frame via the extrinsic matrix $[\bm{R} | \bm{t}]$, representing a rigid body transformation: $\bm{P}^{cam}_i = \bm{R} \bm{P}_i + \bm{t}$. The transformed 3D points are projected into 2D pixel coordinates $(u_i, v_i)$ through the camera intrinsic matrix $\bm{K}$. 
These coordinates are used as attention anchors to sample features from the multi-scale feature maps $\mathcal{F}$.

% \textbf{Point-wise Feature Sampling\,} Using the attention anchors $(u_i, v_i)_{i=0}^s$, we extract point-wise visual features from the multi-scale feature maps $\mathcal{F}$. For each point index $i\in\{0, ..., s\}$ and feature map index $n\in\{1, ..., 5\}$, we sample point-wise feature vectors $f_{i,n}$ using bilinear interpolation: $\bm{f}_{i,n} = \text{BilinearInterp}(\mathcal{F}_n, u_i, v_i) \in \mathbb{R}^d$, 
% where $d$ is the channel dimension of the multi-scale feature maps.
% We then aggregate feature vectors at multiple resolutions and trajectory points into a single spatial context vector $\bm{F}$. Here, we apply average pooling only along the temporal axis and preserve multi-resolution information by simply concatenating the pooled feature vectors $\bar{\bm{f_n}}$. Specifically, the spatial context vector $\bm{F}$ is derived as
% $$\bm{F} = [\bar{\bm{f}}_1, \dots, \bar{\bm{f}}_5] \in \mathbb{R}^{5d}, \quad \text{where  } \bar{\bm{f}}_n = \frac{1}{s+1} \sum_{i=0}^s \bm{f}_{i,n}.$$
% This yields a compact spatial feature that preserves multi-scale visual cues around the attention anchors.

\textbf{Point-wise Feature Sampling.\,} Using the attention anchors $(u_i, v_i)_{i=0}^s$, we extract point-wise visual features from the multi-scale feature maps $\mathcal{F}$. For each point index $i\in\{0, ..., s\}$ and feature map index $n\in\{1, ..., 5\}$, we sample point-wise feature vectors $\bm{f}_{i,n}$ using bilinear interpolation:
$\bm{f}_{i,n} = \text{BilinearInterp}(\mathcal{F}_n, u_i, v_i) \in \mathbb{R}^d$,
where $d$ is the channel dimension of the multi-scale feature maps.
We then aggregate feature vectors across multiple resolutions and trajectory points into a single spatial context vector $\bm{F}$. Here, we apply average pooling only along the temporal axis while preserving multi-resolution information through concatenation of the pooled feature vectors $\bar{\bm{f}}_n$. Specifically, the spatial context vector $\bm{F}$ is computed as follows, yielding a compact feature representation that preserves multi-scale visual information centered around the attention anchors.
\begin{equation}
    \bm{F} = [\bar{\bm{f}}_1, \dots, \bar{\bm{f}}_5] \in \mathbb{R}^{5d},
    \quad
    \text{where }
    \bar{\bm{f}}_n
    =
    \frac{1}{s+1}
    \sum_{i=0}^s
    \bm{f}_{i,n}.
    \label{eq:spatial_context}
\end{equation}

% Specifically, the spatial context vector $\bm{F}$ is derived as
% \begin{equation}
%     \bm{F} = [\bar{\bm{f}}_1, \dots, \bar{\bm{f}}_5] \in \mathbb{R}^{5d},
%     \quad
%     \text{where }
%     \bar{\bm{f}}_n
%     =
%     \frac{1}{s+1}
%     \sum_{i=0}^s
%     \bm{f}_{i,n}.
% \end{equation}
% This yields a compact feature representation that preserves multi-scale visual information centered around the attention anchors.

\vspace{-5mm}

\subsection{Diffusion Process}
\label{subsec:diffusion_process}
% We formalize our policy as a conditional diffusion process following Diffusion Policy (DP) framework~\citep{chi2023diffusionpolicy}, using the spatial context $\bm{F}_k$ as the conditioning signal instead of the fixed visual observation $\mathcal{O}_t$. The training loss is defined as:

We formulate our policy as a conditional diffusion process following the Diffusion Policy (DP) framework~\citep{chi2023diffusionpolicy}, where the denoising network is conditioned on the spatial context $\bm{F}_k$ instead of the fixed visual observation $\bm{O}$. The training loss is defined as:
\[
\mathcal{L} = \mathbb{E}_{k, \bm{A}_0, \bm{\epsilon}} \left[
\left\| \bm{\epsilon} -
\bm{\epsilon}_{\theta}(\bm{A}_k, \bm{F}_k, \bm{q}, k)
\right\|^2
\right],
\]
where $\bm{A}_k = \sqrt{\bar{\alpha}_k} \bm{A}_0 + \sqrt{1-\bar{\alpha}_k} \bm{\epsilon}$ is the perturbed action at diffusion step $k$ with $\bm{\epsilon} \sim \mathcal{N}(0, \bm{I})$, and $\bm{q}$ denotes the observed robot state.
%The action denoising network is built upon a conditional U-Net backbone, where the spatial context $\bm{F}_k$ is incorporated via Feature-wise Linear Modulation (FiLM)~\citep{perez2018film}. 
The action denoising network uses a conditional U-Net backbone with Feature-wise Linear Modulation (FiLM)~\citep{perez2018film}.
During inference, we employ the DDIM scheduler~\citep{song2020denoising} to enable high-quality action generation with fewer denoising steps.

\section{Simulation Experiments}
\label{sec:simulation}
We conduct three sets of comparative experiments. 1) We evaluate the general performance of SCDP in single-view settings across a diverse range of manipulation tasks. 2) We focus on challenging single-view manipulation tasks and assess the effectiveness of spatial conditioning against methods that explicitly model task relevance. 3) We compare SCDP with other baselines under various camera configurations to demonstrate its effectiveness in single-view settings.

% \begin{enumerate}
%     \item We evaluate the general performance of SCDP in single-view settings across a diverse range of manipulation tasks.
%     \item We focus on challenging single-view manipulation tasks and assess the effectiveness of spatial conditioning against methods that explicitly model task relevance.
%     \item We compare SCDP with other baselines under various camera configurations to demonstrate its effectiveness in single-view settings.
% \end{enumerate}

% We evaluate SCDP on a comprehensive suite of 54 tasks across Meta-World~\citep{yu2020meta} and DexArt~\citep{bao2023dexart} to assess its performance across diverse manipulation scenarios. 

% We evaluate SCDP on Meta-World~\citep{yu2020meta} and DexArt~\citep{bao2023dexart} benchmarks.
% Meta-World consists of diverse manipulation tasks performed with a parallel gripper. Its tasks are grouped by difficulty following the protocol of~\citet{seo2023masked} (see Appendix~\ref{appendix:meta-world_difficulty} for details). DexArt focuses on dexterous manipulation of articulated objects using an Allegro hand, providing a complementary benchmark with more complex hand-object interactions. 
% For the first experiment, we used all suite of 54 tasks across Meta-World and DexArt evaluated in single-camera setup. For the second and third experiment, we evaluated on Meta-World Hard suite in the same single-camera setting. 
\subsection{Experimental Setup}

\textbf{Simulation Benchmarks.} We evaluate SCDP on the Meta-World~\citep{yu2020meta} and DexArt~\citep{bao2023dexart} benchmarks. 
Meta-World consists of diverse parallel-gripper manipulation tasks, grouped by difficulty following~\citet{seo2023masked} (see Appendix~\ref{appendix:meta-world_difficulty} for details). 
DexArt provides a complementary benchmark for dexterous manipulation of articulated objects with an Allegro hand, involving more complex hand-object interactions. For the first experiment, we use the full suite of 54 tasks across Meta-World and DexArt. For the second and third experiments, we focus on the Meta-World Hard suite.

% \textbf{Training Details.} All experiments are conducted across three random seeds $\{0, 1, 2\}$. For Meta-World, models are trained for 1,000 epochs using 20 scripted expert demonstrations, whereas for DexArt, models are trained for 3,000 epochs using 100 RL-expert demonstrations. We report the best-checkpoint success rate for Meta-World and the top-three checkpoint average for DexArt.

\textbf{Training Details.} All experiments are conducted across three random seeds $\{0, 1, 2\}$. For Meta-World, models are trained for 1,000 epochs using 20 scripted expert demonstrations, whereas for DexArt, models are trained for 3,000 epochs using 100 RL-expert demonstrations. Checkpoints are saved and evaluated every 200 epochs. We report the best-checkpoint success rate for Meta-World and the top-three checkpoint average for DexArt, corresponding to the top 20\% in both cases.

% \textbf{Baselines.}
% We compare SCDP with baselines including both general visual conditioning methods and recent methods that explicitly extract task-relevant visual information. DP~\citep{chi2023diffusionpolicy} and DP3~\citep{ze20243d} perform diffusion-based action denoising conditioned on 2D image features and 3D point-cloud features, respectively. SKIL~\citep{wang2025skil} extracts semantic keypoints using vision foundation models, and OTTER~\citep{huang2025otter} selects text-aware task-relevant visual features for action prediction.

% We compare SCDP with baselines selected according to the goals of each experiment. For Experiments~1 and~3, we use DP~\citep{chi2023diffusionpolicy} and DP3~\citep{ze20243d} as representative 2D image- and 3D point-cloud-based diffusion policies. For Experiment~2, we compare against SKIL~\citep{wang2025skil} and OTTER~\citep{huang2025otter}, which explicitly extract task-relevant visual information.
\textbf{Baselines.} We compare SCDP with baselines selected according to the goals of each experiment. 
% For Experiment~1, we use DP~\citep{chi2023diffusionpolicy} and DP3~\citep{ze20243d} as representative 2D image- and 3D point-cloud-based diffusion policies. 
For Experiment~1, we compare SCDP with DP~\citep{chi2023diffusionpolicy}, a representative imitation learning baseline, and DP3~\citep{ze20243d}, its point-cloud-based variant.
For Experiment~2, we compare against SKIL~\citep{wang2025skil} and OTTER~\citep{huang2025otter}, which explicitly extract task-relevant visual information, with DP serving as a general visual conditioning baseline. For Experiment~3, we compare against DP with an additional wrist camera and DP3 to evaluate SCDP's competitiveness under a single-RGB setting.

% Main Simulation Results
\begin{table}[!t]
\centering
\caption{\textbf{General Performance across Diverse Manipulation Tasks.} Success rates (\%) on Meta-World and DexArt benchmarks. Numbers in parentheses indicate the number of tasks.}
\label{tab:main_simulation_results}
\resizebox{0.8\textwidth}{!}{
\begin{tabular}{lcccc|c|c}
\toprule
\diagbox{Method}{Suite} 
& \begin{tabular}[c]{@{}c@{}}Meta-World\\ Easy (28) \end{tabular} & \begin{tabular}[c]{@{}c@{}}Meta-World\\ Medium (11)\end{tabular} & \begin{tabular}[c]{@{}c@{}}Meta-World\\ Hard (6)\end{tabular} & \begin{tabular}[c]{@{}c@{}}Meta-World\\ Very Hard (5)\end{tabular} & \begin{tabular}[c]{@{}c@{}}DexArt\\ (4) \end{tabular} & \begin{tabular}[c]{@{}c@{}} Average\\ (54) \end{tabular} \\ \midrule
DP & 79.9 & 36.1 & 16.7 & 46.0 & 56.8 & 59.1 \\ 
DP3 & \textbf{92.3} & 78.5 & 60.3 & 82.7 & 52.6 & 82.1 \\
\rowcolor{gray!20}
\textbf{SCDP (Ours)} & 91.5 & \textbf{83.0} & \textbf{82.5} & \textbf{89.7} & \textbf{62.4} & \textbf{86.5}\\
\bottomrule
\end{tabular}}
\vspace{-3mm}
\end{table}

% Table 2
% DP, SKIL, Otter, SCDP
\begin{table}[t]
\centering
\caption{\textbf{Comparison with Task-Relevant Visual Conditioning Methods.}
Success rates (\%) are reported for each task. The average and standard deviation are computed over the Hard tasks only.}

\label{tab:task_relevant}

\resizebox{\textwidth}{!}{%
\begin{tabular}{lcc!{\vrule width 0.4pt}*{6}{c}c}
\toprule
\multirow{2}{*}{Method}
& Easy & Medium & \multicolumn{6}{c}{Hard}
& \multirow{2}{*}{Average $\pm$ STD} \\
\cmidrule(lr){2-9}
& Button-press & Hammer & Assembly & Hand-insert & Pick-out-of-hole & Pick-place & Push & Push-back & \\ 

\midrule
DP & 100.0 & 35.0 & 15.0 & 16.7 & 11.7 & 8.3 & 20.0 & 28.3 & 16.7 $\pm$ 7.0 \\
SKIL & 100.0 & 100.0 & 85.0 & 11.7 & 1.7 & 13.3 & 51.7 & 1.7 & 27.5 $\pm$ 33.7 \\
OTTER & 98.3 & 88.3 & 70.0 & 43.3 & 60.0 & 38.3 & 33.3 & 38.3 & 47.2 $\pm$ 14.5 \\
\rowcolor{gray!20}
\textbf{SCDP (Ours)} & 100.0 & 100.0 & \textbf{91.7} & \textbf{81.7} & \textbf{80.0} & \textbf{86.7} & \textbf{93.3} & \textbf{61.7} & \textbf{82.5 $\pm$ 11.5} \\

\bottomrule
\end{tabular}%
}
\vspace{-3mm}
\end{table}

% % Table 3
% DP w/ Wrist, DP3, SCDP
\begin{table}[t]
\centering
\caption{\textbf{Comparison across Camera Configurations.}
Success rates (\%) are reported for each Meta-World ``Hard'' task, with average and standard deviation computed over all tasks.}
\label{tab:setting}

\resizebox{\textwidth}{!}{%
\begin{tabular}{ll cccccc c}
\toprule
\multirow{2}{*}{Method} & \multirow{2}{*}{\makecell{Camera\\Setting}}
& \multicolumn{6}{c}{Hard}
& \multirow{2}{*}{Average $\pm$ STD} \\
\cmidrule(lr){3-8}
& & Assembly & Hand-insert & Pick-out-of-hole & Pick-place & Push & Push-back & \\ 
\midrule

DP & w/ Wrist & 95.0 & \textbf{98.3} & 20.0 & 73.3 & 91.7 & \textbf{66.7} & 74.2 $\pm$ 29.4 \\
DP3 & w/ Depth & \textbf{100.0} & 28.3 & 31.7 & 85.0 & \textbf{98.3} & 18.3 & 60.3 $\pm$ 38.0 \\
\rowcolor{gray!20}
\textbf{SCDP (Ours)} & Single RGB & 91.7 & 81.7 & \textbf{80.0} & \textbf{86.7} & 93.3 & 61.7 & \textbf{82.5 $\pm$ 11.5} \\
\bottomrule
\end{tabular}%
}
\vspace{-2mm}
\end{table}

% % More Baselines
% \begin{table}[t]
% \centering
% \caption{\textbf{Comparison with more baselines on Meta-World Hard tasks.} Success rates (\%) are reported along with the average and standard deviation (STD).}
% \label{tab:more_baselines}
% \resizebox{\textwidth}{!}{%
% \begin{tabular}{lccccccc}
% \toprule
% \diagbox{Method}{Task} & Assembly & Hand-insert & Pick-out-of-hole & Pick-place & Push & Push-back & \textbf{Average $\pm$ STD} \\ \midrule
% SCDP (Ours) & 91.7 & \textbf{81.7} & \textbf{80.0} & \textbf{86.7} & 93.3 & \textbf{61.7} & \textbf{82.5 $\pm$ 11.5} \\
% DP & 15.0 & 16.7 & 11.7 & 8.3 & 20.0 & 28.3 & 16.7 $\pm$ 7.0 \\
% ACT & 20.0 & 33.3 & 13.3 & 30.0 & 51.7 & 5.0 & 25.6 $\pm$ 16.5 \\
% DP3 & \textbf{100.0} & 28.3 & 31.7 & 85.0 & \textbf{98.3} & 18.3 & 60.3 $\pm$ 38.0 \\
% Otter & 70.0 & 43.3 & 60.0 & 38.3 & 33.3 & 38.3 & 47.2 $\pm$ 14.5 \\
% SKIL & 85.0 & 11.7 & 1.7 & 13.3 & 51.7 & 1.7 & 27.5 $\pm$ 33.7 \\ \bottomrule
% \end{tabular}%
% }
% \vspace{-5mm}
% \end{table}

% Data Scaling Plot
\begin{figure}[!t]
    \centering
    \includegraphics[width=0.8\linewidth]{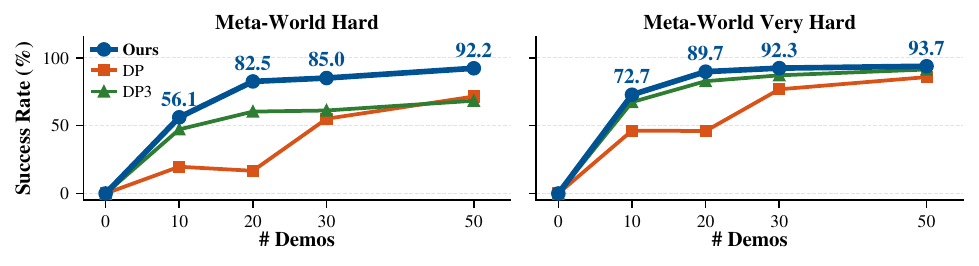}
    \caption{\textbf{Data Efficiency and Scaling.} Success rates of SCDP and baselines with varying numbers of demonstrations on Meta-World ``Hard'' (6 tasks) and ``Very Hard'' (5 tasks).}
    \label{fig:scaling}
    \vspace{-5mm}
\end{figure}

% Ablation Plot
\begin{figure}[t] 
    \centering
    \includegraphics[width=\linewidth]{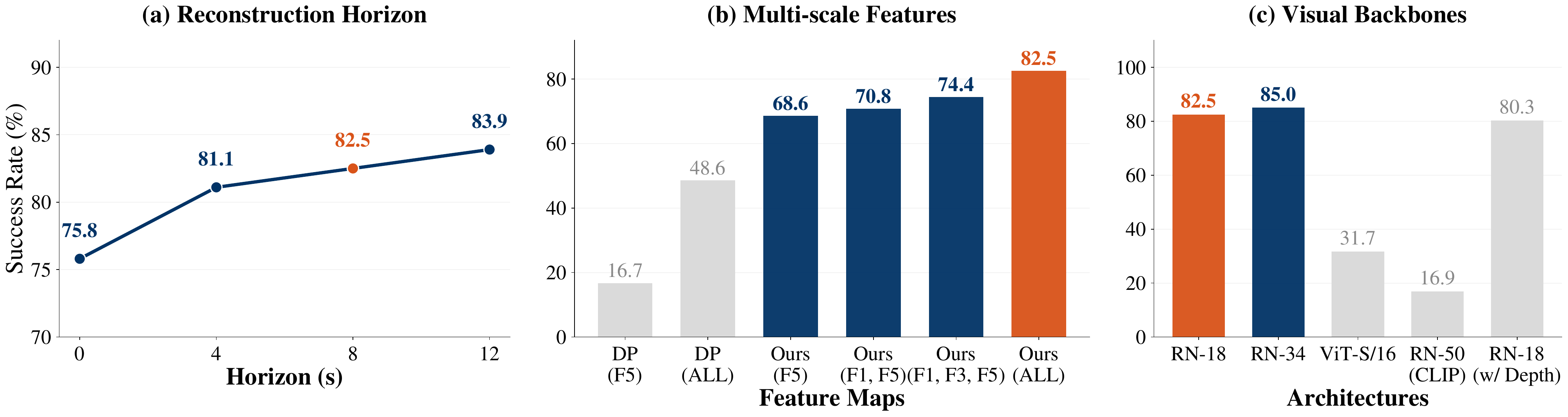}
    % \caption{\textbf{Ablation Study.} Effects of (a) reconstruction horizon, (b) multi-scale features, and (c) visual backbones. Orange markers indicate the default configuration of our final model.}
    % \caption{\textbf{Ablations} on the effects of (a) reconstruction horizon, (b) multi-scale features, and (c) visual backbones. Orange markers indicate the default configuration of our final model.}
    \caption{\textbf{Design Ablations.} Effects of (a) reconstruction horizon, (b) multi-scale features, and (c) visual backbones. Orange markers indicate the default configuration of our final model.}
    \label{fig:ablation}
\end{figure}

\subsection{Simulation Results}
% \subsection{Performance, Robustness, and Efficiency}

\textbf{Strong Performance on Challenging Manipulation Tasks.}
Experiment~1 shows that SCDP achieves strong overall performance across diverse manipulation tasks, with particularly large gains on challenging tasks.
Table~\ref{tab:main_simulation_results} reports the average success rates across Meta-World difficulty groups and DexArt tasks. SCDP achieves over 80\% success in every Meta-World difficulty group using only 20 demonstrations and a single global camera.
Notably, SCDP outperforms the baselines by larger margins on the ``Hard'' and ``Very Hard'' groups, empirically demonstrating the effectiveness of our approach on challenging manipulation tasks. SCDP also achieves the highest success rate on DexArt tasks, which require controlling a high-DoF Allegro hand and interacting with articulated objects. This result further supports the applicability of spatial conditioning across diverse manipulation settings. Detailed results are reported in Appendix~\ref{appendix:simulation_results}.

\textbf{Effectiveness of Spatial Conditioning.}
Experiment~2 demonstrates that spatial conditioning helps SCDP capture task-relevant visual information in challenging single-view manipulation tasks.
Table~\ref{tab:task_relevant} compares SCDP with baselines that explicitly incorporate task-relevant visual cues.
DP, which relies on a global scene embedding, shows the lowest performance, while SKIL and OTTER achieve higher success rates, suggesting the benefit of spatially selective visual conditioning.
SCDP outperforms all baselines by a large margin while maintaining a relatively low standard deviation, indicating that spatial conditioning provides effective and consistent representations.

%captures task-relevant information even in the single-view setting.

% We further evaluate SCDP on the Meta-World ``Hard'' tasks, which consist of six precise manipulation tasks, to analyze the effectiveness of the proposed spatial conditioning mechanism. Specifically, we compare SCDP against: 1) baselines using an additional wrist-mounted camera, and 2) single-camera baselines, including methods that explicitly extract task-relevant features using pretrained modules.

% \textbf{Single RGB View Competitiveness.}
% Experiment 3 compares different camera configurations on Meta-World ``Hard'' tasks to evaluate their performance in complex contact and precise manipulation scenarios. As shown in Table~\ref{tab:setting}, despite using only a single RGB camera, SCDP achieves performance comparable to DP with an additional wrist-view camera and DP3 using depth information, while attaining the highest average success rate and the lowest standard deviation.

\textbf{Single RGB View Competitiveness.}
Experiment~3 shows that SCDP remains competitive even when using only a single RGB camera.
As shown in Table~\ref{tab:setting}, SCDP achieves performance comparable to DP with an additional wrist-view camera and DP3 with depth observations on Meta-World ``Hard'' tasks, while attaining the highest average success rate and the lowest standard deviation.

%Table~\ref{tab:more_baselines} further shows that SCDP outperforms SKIL and Otter, which use explicit task-relevant feature extraction mechanisms, demonstrating that SCDP learns effective representations without requiring an external pretrained module.

% \textbf{Robustness across challenging tasks.}
% \textbf{Precise and Challenging Tasks.}
% Table~\ref{tab:more_baselines} compares all baselines on the two basic Meta-World tasks and ``Hard'' tasks, which involve complex contacts and precise manipulation. %where reliably capturing critical local cues strongly affects overall task success. 
% We further compare SCDP with wrist-camera-based methods to assess whether a single world-camera policy can achieve comparable performance without relying on wrist-mounted views.

% Existing methods, including explicit task-relevant feature extraction approaches, show high variance and substantial performance drops on certain tasks. 
% In contrast, SCDP achieves the highest average success rate of 82.5\% while maintaining a relatively low standard deviation. 
% This suggests that spatial conditioning provides more informative and consistent representations across diverse manipulation tasks.

\subsection{Ablation Study}
Ablation studies are conducted on the Meta-World ``Hard'' group unless otherwise specified.
%We conduct all ablation studies on the Meta-World ``Hard'' group.%, which consists of six tasks requiring complex contacts and precise manipulation.

\textbf{Data Efficiency.}
Spatial conditioning provides a useful inductive bias toward task-relevant regions, which we expect to enable more data-efficient learning. As illustrated in Fig.~\ref{fig:scaling}, SCDP exhibits superior data efficiency, supporting the effectiveness of its visual representations. As the number of demonstrations increases, SCDP continues to improve steadily and reaches a higher performance ceiling, whereas DP and DP3 show slower convergence and lower overall performance. These results suggest that SCDP can learn more effectively by leveraging task-relevant visual features.

\textbf{Effects of Multi-Scale Image Encoder.} Fig.~\ref{fig:ablation}(b) summarizes the impact of integrating various feature map levels. Comparing an SCDP variant using only high-level semantic features (F5) against multi-scale configurations, we observe a clear upward trend in success rates. Peak performance reaches 82.5\% when all scales (F1--F5) are utilized, representing a 13.9\%p improvement over the F5-only variant. Notably, even DP benefits substantially from using all feature levels, further supporting the broad utility of multi-scale visual information for precise action generation.

\textbf{Effects of Reconstruction Horizon.} Fig.~\ref{fig:ablation}(a) illustrates the impact of the reconstruction horizon~$s$. Success rates improve consistently with the horizon, rising from 75.8\% at $s=0$ (baseline with a single static end-effector point) to 83.9\% at $s=12$. This gain suggests that incorporating future trajectory points provides richer task-relevant context for action generation, leading to a more effective representation. We select \textbf{$s=8$} (82.5\%) as the default for all experiments, as it achieves near-optimal performance with fewer trajectory queries than $s=12$. In Fig.~\ref{fig:ablation}(b), DP (F5) and Ours (F5) differ mainly in spatial conditioning, and the over 40\%p gap further confirms its effectiveness.

\textbf{Choice of Visual Backbone.} Fig.~\ref{fig:ablation}(c) compares various architectures as SCDP's feature extractor. ViT-S/16 (31.7\%) and CLIP-pretrained ResNet-50 (16.9\%) underperform significantly, likely due to ViT's weaker locality bias under limited data and CLIP's high-level semantic bias.
In contrast, standard ResNets better preserve locality bias across resolutions.
While ResNet-34 achieves the highest success rate (85.0\%), we adopt ResNet-18 (82.5\%) as our primary backbone for a fair comparison. The lack of improvement from adding depth further supports our RGB-based design.

\begin{figure}[t] 
    \centering
    \includegraphics[width=1.0\linewidth]{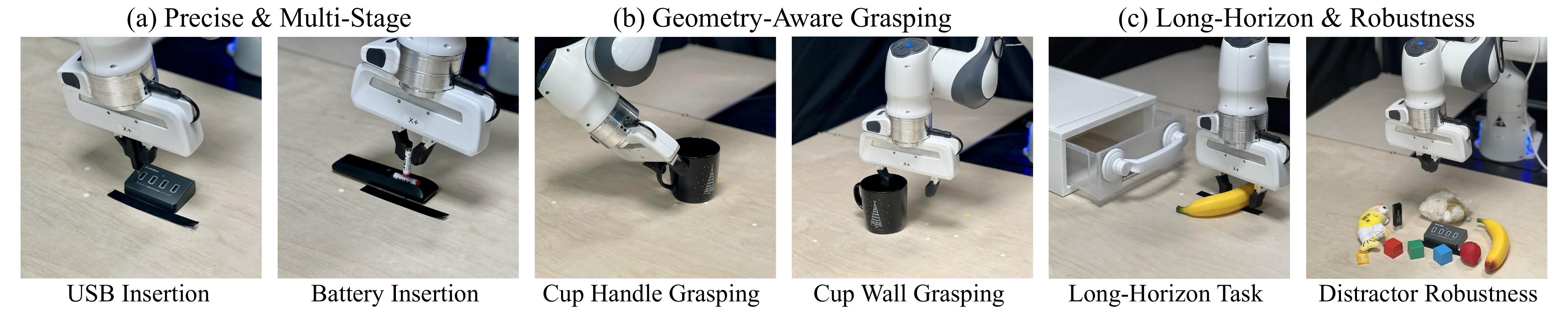}
    \caption{\textbf{Real-World Experiments.} Six real-world tasks in a single-camera setting.}
    \label{fig:real_experiment}
    \vspace{-5mm}
\end{figure}

% Real World Experiment Results
\begin{table*}[t]
\centering
\caption{\textbf{Real-World Experiment Results.} All experiments are conducted using a single global-view camera. Success rates (\%) are reported over 20 trials.}
\label{tab:real_experiment}

\setlength{\tabcolsep}{3pt}
\resizebox{0.8\textwidth}{!}{
\begin{tabular}{l ccc ccc cc cc}
\toprule
& \multicolumn{3}{c}{\textbf{USB Insertion}} 
& \multicolumn{3}{c}{\textbf{Battery Insertion}} 
& \multicolumn{2}{c}{\textbf{Cup Handle Grasping}} 
& \multicolumn{2}{c}{\textbf{Cup Wall Grasping}} \\
\cmidrule(lr){2-4} 
\cmidrule(lr){5-7} 
\cmidrule(lr){8-9} 
\cmidrule(lr){10-11}
& Grasp & Insert & Push 
& Grasp & Insert & Push 
& Original & Distractor
& Original & Distractor \\
\midrule

DP & 60 & 0 & 0 
& 90 & 40 & 20 
& 50 & 30 
& 80 & 65 \\

% ACT & 40 & 0 & 0 
% & 20 & 0 & 0 
% & 45 & 20 
% & 55 & 35 \\

DP3 & 70 & 0 & 0 
& 90 & 40 & 5 
& 75 & 25 
& 80 & 40 \\

OTTER & 55 & 0 & 0 
& 65 & 25 & 5 
& 45 & 35 
& 75 & 50 \\

SKIL & 80 & 10 & 0 
& 80 & 20 & 5 
& 70 & 55 
& 85 & 75 \\

\rowcolor{gray!20}
\textbf{SCDP (Ours)} & \textbf{85} & \textbf{40} & \textbf{30} 
& \textbf{100} & \textbf{70} & \textbf{55} 
& \textbf{95} & \textbf{85} 
& \textbf{95} & \textbf{80} \\

\bottomrule
\end{tabular}
}

\setlength{\tabcolsep}{3pt}
\resizebox{0.8\textwidth}{!}{
\begin{tabular}{l cccc ccc c}
\toprule
& \multicolumn{4}{c}{\textbf{Long Horizon}} 
& \multicolumn{3}{c}{\textbf{Push Cube}} 
& \textbf{Average} \\
\cmidrule(lr){2-5} 
\cmidrule(lr){6-8} 
\cmidrule(lr){9-9}
& \makecell{Open\\Drawer} 
& \makecell{Pick\\Banana} 
& \makecell{Place\\Banana} 
& \makecell{Close\\Drawer} 
& Original & Distractor & Cluttered
& Success Rate (\%) \\
\midrule

DP 
& 85 & 70 & 45 & 30 
& \textbf{90} & 50 & 50 
& 50.3 \\

% ACT 
% & 85 & 10 & 0 & 0 
% & 75 & 55 & 10 
% & 26.5 \\

DP3 
& 85 & 30 & 0 & 0 
& 40 & 20 & 0 
& 35.3 \\

OTTER 
& 80 & 45 & 5 & 5 
& 45 & 0 & 5 
& 31.8 \\

SKIL 
& 75 & 35 & 35 & 25 
& 50 & 10 & 25 
& 43.2 \\

\rowcolor{gray!20}
\textbf{SCDP (Ours)} 
& \textbf{90} & \textbf{75} & \textbf{65} & \textbf{60} 
& 85 & \textbf{75} & \textbf{70} 
& \textbf{73.8} \\

\bottomrule
\end{tabular}
}
\vspace{-5mm}
\end{table*}

\section{Real-World Experiments}
\label{sec:real_world}

\subsection{Task Definitions}
\begin{description}[leftmargin=0pt, itemsep=2pt]
\item[\textbf{USB Insertion.}] A high-precision task:
(1) \textit{Grasp}: picking a USB drive;
(2) \textit{Insert}: aligning and partially inserting the connector into a hub port;
(3) \textit{Push}: fully seating the USB into the port. This stage requires precise geometric alignment to apply force along the connector axis.

\item[\textbf{Battery Insertion.}] A precise and contact-rich task:
(1) \textit{Grasp}: picking a battery;
(2) \textit{Insert}: placing the battery into a slot;
(3) \textit{Push}: overcoming spring resistance to securely seat the battery.

\item[\textbf{Cup Handle Grasping.}]
A task requiring geometry-aware grasping of the cup handle. % under rotation variations of $\pm 60^{\circ}$. 
The policy must infer the handle orientation and adapt its grasp accordingly.

\item[\textbf{Cup Wall Grasping.}]
A task requiring precise grasping of the cup wall. % under randomized cup poses. 
The policy must insert one gripper finger into the cup while avoiding collisions with the handle and cup rim.

\item[\textbf{Long Horizon.}]
(1) \textit{Open Drawer}: opening a drawer; % from a randomized position;
(2) \textit{Pick Banana}: grasping a banana; % with randomized pose;
(3) \textit{Place Banana}: placing the banana into the drawer; 
(4) \textit{Close Drawer}: closing the drawer.

%\item[\textbf{Push Cube.}] Pushing the red cube. % with randomized pose;
\item[\textbf{Push Cube.}] Pushing the red cube to evaluate robustness to unseen distractors and dense clutter.
\end{description}

The evaluation settings are defined as follows:
\textit{Original}, the original environment;
\textit{Distractor}, with 4--7 unseen distractor objects; and
\textit{Cluttered}, with 4--7 seen objects arranged in dense clutter.

% \begin{itemize}[leftmargin=15pt, nosep]
% \item \textit{Original}: Evaluation in the original setting.
% \item \textit{Distractor}: Testing robustness with 4--7 unseen distractor objects.
% \item \textit{Cluttered}: Testing robustness in a high-density environment with 4--7 seen objects.
% \end{itemize}

\subsection{Real-World Experiment Results}
Table~\ref{tab:real_experiment} summarizes the success rates across real-world tasks (see Appendix~\ref{appendix:realworld_setup} for setup details).

\textbf{High-Precision Manipulation.}
The \textit{USB} and \textit{Battery Insertion} tasks require precise control under narrow tolerances and contact-rich interactions. SCDP performs these challenging tasks using only a single global-view camera by leveraging the reconstructed end-effector trajectory as attention anchors. This enables SCDP to focus on anticipated interaction regions rather than the entire scene. Combined with multi-scale visual features that preserve fine-grained local details, SCDP effectively captures local information around contact regions. %In contrast, the baselines often fail to localize these critical regions, leading to near-zero success rates.

% \textbf{Geometry-Aware Grasping.}
% The cup grasping tasks require identifying feasible grasp regions among visually similar local structures, such as the handle, cup wall, and interior. SCDP extracts visual features along the predicted action trajectory, allocating representational capacity to interaction-relevant regions and thereby more efficiently distinguishing subtle geometric cues.

\textbf{Distractor Robustness.}
The two cup grasping tasks in distractor settings require identifying task-relevant regions from unstructured scenes. Since SCDP focuses on regions along the predicted end-effector trajectory, such as the cup handle and cup wall during approach, it reduces sensitivity to surrounding clutter. The \textit{Push Cube} experiment further demonstrates SCDP's robustness to environmental variations. While DP's performance drops from 90\% to 50\% in both \textit{Distractor} and \textit{Cluttered} scenes, SCDP maintains relatively high success rates of 75\% and 70\%, respectively. These results suggest that grounding visual conditioning along the estimated end-effector trajectory enables SCDP to focus on task-relevant regions while remaining robust to surrounding clutter.

%\textbf{Long-Horizon Reliability.}
% The four-stage Long-Horizon task shows that, although SCDP extracts local visual regions, it can preserve task-level context through the multi-scale image encoder. This enables SCDP to handle sequential manipulation without losing global task structure.

% \textbf{Robustness to Visual Clutter.}
% The Push Cube experiment highlights SCDP's resilience to environmental variations. While DP's performance drops from 90\% to 50\% in both \textit{Distractor} and \textit{Cluttered} scenes, SCDP maintains high success rates of 75\% and 70\%, respectively. The distractor settings in the two cup grasping tasks show a similar trend, further supporting SCDP's robustness. These results suggest that grounding visual conditioning along the estimated end-effector trajectory helps SCDP emphasize task-relevant regions while reducing sensitivity to surrounding clutter.

%===============================================================================

\section{Conclusion and Limitations}
\label{sec:conclusion}
    We presented SCDP for precise and robust manipulation using a single RGB camera. 
    By using intermediate action trajectories as spatial queries over multi-scale visual features, SCDP attends to task-relevant regions while capturing both global context and fine-grained geometry. 
    Experiments in simulation and real-world tasks demonstrate its strong performance, efficiency, and robustness. 
    Notably, SCDP achieves performance comparable to a baseline equipped with an additional wrist-view camera, while relying solely on a single RGB camera. This result highlights the potential of single-camera learning as a practical and scalable alternative for robot learning.

    Although SCDP is effective, it assumes that task-relevant visual information lies near the end-effector trajectory. This assumption may limit its applicability to tool-use or indirect-interaction tasks, where task-relevant contacts occur away from the end-effector. In addition, in heavily occluded settings, extracting reliable visual features from the attention region becomes challenging. Future work could address these limitations by extending the attention region beyond the end-effector trajectory and leveraging complementary sensing modalities.

    % Submission to CoRL 2026 will be entirely electronic, via a web site (not email). Information about the submission process and \LaTeX{} templates are available on the conference web site at \url{https://corl.org/}. For camera ready submission, use the \texttt{final} option for the \texttt{\textbackslash usepackage} command.
    
% \clearpage
%===============================================================================
	
%     Submission to CoRL 2026 will be entirely electronic, via a web site (not email). Information about the submission process and \LaTeX{} templates are available on the conference web site at \url{https://corl.org/}. For camera ready submission, use the \texttt{final} option for the \texttt{\textbackslash usepackage} command. 

%===============================================================================

\clearpage
% The acknowledgments are automatically included only in the final and preprint versions of the paper.
% \acknowledgments{If a paper is accepted, the final camera-ready version will (and probably should) include acknowledgments. All acknowledgments go at the end of the paper, including thanks to reviewers who gave useful comments, to colleagues who contributed to the ideas, and to funding agencies and corporate sponsors that provided financial support.}

\acknowledgments{
This work was supported by the National Research Foundation of Korea (NRF) grant funded by the Korean government (MSIT) (RS-2026-25615611). This research has been funded by the Industrial Technology Innovation Program (P0028404) of the Ministry of Industry, Trade and Energy of Korea. This work was also supported by the National Research Foundation of Korea (NRF) grant funded by the Korean government (MSIT) in 2026 (Grant No. RS-2025-02213804, Project Title: In-Space Servicing and Manufacturing Research Center). 
}

%===============================================================================

\bibliography{example}  % .bib

\clearpage
\appendix
\section{Implementation Details}
\label{appendix:implementation_details}

\subsection{Training Hyperparameters}

We summarize the implementation details and training hyperparameters used for SCDP in Table~\ref{tab:scdp_implementation_details}.

\begin{table}[h]
\centering
\caption{Implementation details and hyperparameters of SCDP.}
\label{tab:scdp_implementation_details}
\resizebox{0.75\textwidth}{!}{
\setlength{\tabcolsep}{6pt}
\begin{tabular}{l l}
\toprule
\textbf{Hyperparameter} & \textbf{Value} \\
\midrule
\multicolumn{2}{c}{\textbf{Training Hyperparameters}} \\
\midrule
Epochs & 1000 (Meta-World); 3000 (Real-World, DexArt) \\
Batch size & 128 \\
Optimizer & Adam \\
Learning rate & \texttt{1e-4} \\
LR scheduler & Cosine \\
Weight decay & \texttt{1e-6} \\
Reconstruction horizon & 8 \\
\midrule
\multicolumn{2}{c}{\textbf{Diffusion Head}} \\
\midrule
U-Net dimensions & $(128, 256, 384)$ \\
Noise scheduler & DDIM \\
Denoising steps & 100 (train); 10 (real-world inference); 16 (simulation inference) \\
Prediction horizon & 16 \\
Observation horizon & 2 \\
Action horizon & 8 \\
\midrule
\multicolumn{2}{c}{\textbf{Multi-Scale Image Encoder}} \\
\midrule
Backbone channels & $(64, 64, 128, 256, 512)$ \\
Projection dimension & 64 \\
Feature levels & F1--F5 \\
Output dimension & 320 \\
\bottomrule
\end{tabular}}
\end{table}

\section{Supplementary Materials: Simulation Experiments}
\label{appendix:simulation_results}

\subsection{Meta-World Benchmark} 
\label{appendix:meta-world_difficulty}
We follow Seo et al.~\citep{seo2023masked} for the Meta-World task grouping, and refer to the LeRobot Meta-World documentation\footnote{\url{https://huggingface.co/docs/lerobot/metaworld}} for the descriptive interpretation of each difficulty group summarized in Table~\ref{tab:metaworld_difficulty_groups}.

\begin{table}[h]
\centering
\caption{Meta-World task difficulty groups used in our simulation experiments.}
\resizebox{0.75\textwidth}{!}{
\label{tab:metaworld_difficulty_groups}
\begin{tabularx}{0.82\linewidth}{lcX}
\toprule
\textbf{Group} & \textbf{\# Tasks} & \textbf{Description} \\
\midrule
Easy      & 28 & Tasks with simple dynamics and single-step goals. \\
Medium    & 11 & Tasks requiring multi-step reasoning. \\
Hard      & 6  & Tasks with complex contacts and precise manipulation. \\
Very Hard & 5  & The most challenging tasks in the suite. \\
\bottomrule
\end{tabularx}}
\end{table}

\subsection{DexArt Benchmark}
We follow the success criteria defined in the original DexArt implementation~\citep{bao2023dexart}.
All evaluations are conducted on seen objects. Each rollout is evaluated for a maximum horizon of 250 steps.

\begin{table}[h]
\centering
\caption{Success criteria for DexArt tasks.}
\label{tab:dexart_success_criteria}
\resizebox{0.75\textwidth}{!}{
\setlength{\tabcolsep}{5pt}
\begin{tabular}{l l l}
\toprule
\textbf{Task} & \textbf{Success description} & \textbf{Criterion} \\
\midrule
Laptop & Grasp the screen and open the lid & \texttt{progress > 0.95} \\
Faucet & Grasp the handle and rotate it by $\sim90^{\circ}$ & \texttt{openness > 1.5} \\
Toilet & Grasp and open the toilet lid & \texttt{progress > 0.95} \\
Bucket & Lift the bucket with a stable form-closure grasp & \texttt{delta\_height > 0.25} \\
\bottomrule
\end{tabular}}
\end{table}

\subsection{Per-Task Results on All 54 Simulation Tasks}

\begin{table}[h]
\centering
\caption{\textbf{Detailed per-task simulation results.}
Values denote mean success rates (\%) across three seeds, with standard deviations shown in parentheses.}
\label{tab:appendix_per_task_results}
\scriptsize
\setlength{\tabcolsep}{2.0pt}
\renewcommand{\arraystretch}{1.15}
\resizebox{\textwidth}{!}{
\begin{tabular}{l ccccccccccc c}
\toprule
 
& & & & & & 
& & & & & & Avg. \\
\midrule

\multicolumn{13}{c}{\textbf{Meta-World Easy}} \\
\cmidrule(lr){1-13}
 & \makecell{button\\-press} 
 & \makecell{button-press\\-topdown} 
 & \makecell{button-press\\-topdown-wall} 
 & \makecell{button\\-press-wall} 
 & \makecell{coffee\\-button} 
 & \makecell{dial\\-turn} 
 & \makecell{door\\-close} 
 & \makecell{door\\-lock} 
 & \makecell{door\\-open} 
 & \makecell{door\\-unlock} 
 & \makecell{drawer\\-close} 
 &  \\
 
\cmidrule(lr){2-12}
DP & \textbf{100.0 (0.0)} & \textbf{100.0 (0.0)} & \textbf{100.0 (0.0)} & \textbf{100.0 (0.0)} & \textbf{100.0 (0.0)} & 21.7 (2.9) & \textbf{100.0 (0.0)} & 86.7 (14.4) & \textbf{100.0 (0.0)} & \textbf{100.0 (0.0)} & \textbf{100.0 (0.0)} &  \\
DP3 & \textbf{100.0 (0.0)} & 98.3 (2.9) & \textbf{100.0 (0.0)} & \textbf{100.0 (0.0)} & 91.7 (14.4) & \textbf{85.0 (18.0)} & \textbf{100.0 (0.0)} & \textbf{100.0 (0.0)} & \textbf{100.0 (0.0)} & \textbf{100.0 (0.0)} & \textbf{100.0 (0.0)} &  \\
\rowcolor{gray!20}
\textbf{SCDP} & \textbf{100.0 (0.0)} & \textbf{100.0 (0.0)} & \textbf{100.0 (0.0)} & \textbf{100.0 (0.0)} & \textbf{100.0 (0.0)} & 46.7 (7.6) & \textbf{100.0 (0.0)} & 91.7 (5.8) & \textbf{100.0 (0.0)} & \textbf{100.0 (0.0)} & \textbf{100.0 (0.0)} & \cellcolor{white} \\

\cmidrule(lr){2-12}
 & \makecell{drawer\\-open} 
 & \makecell{faucet\\-close} 
 & \makecell{faucet\\-open} 
 & \makecell{handle\\-press} 
 & \makecell{handle\\-press-side} 
 & \makecell{handle\\-pull} 
 & \makecell{handle\\-pull-side} 
 & \makecell{lever\\-pull} 
 & \makecell{plate\\-slide} 
 & \makecell{plate\\-slide-back} 
 & \makecell{plate-slide\\-back-side} 
 &  \\
 
\cmidrule(lr){2-12}
DP & 81.7 (27.5) & 90.0 (13.2) & 93.3 (2.9) & \textbf{100.0 (0.0)} & 68.3 (5.8) & 36.7 (24.7) & 6.7 (2.9) & 61.7 (7.6) & \textbf{100.0 (0.0)} & \textbf{100.0 (0.0)} & \textbf{100.0 (0.0)} &  \\
DP3 & \textbf{100.0 (0.0)} & 96.7 (5.8) & \textbf{100.0 (0.0)} & 91.7 (5.8) & 60.0 (8.7) & \textbf{88.3 (2.9)} & \textbf{88.3 (12.6)} & \textbf{91.7 (5.8)} & \textbf{100.0 (0.0)} & \textbf{100.0 (0.0)} & \textbf{100.0 (0.0)} &  \\
\rowcolor{gray!20}
\textbf{SCDP} & \textbf{100.0 (0.0)} & \textbf{100.0 (0.0)} & \textbf{100.0 (0.0)} & \textbf{100.0 (0.0)} & \textbf{100.0 (0.0)} & 83.3 (2.9) & 70.0 (5.0) & 63.3 (5.8) & \textbf{100.0 (0.0)} & \textbf{100.0 (0.0)} & \textbf{100.0 (0.0)} & \cellcolor{white} \\

\cmidrule(lr){2-12}
 & \makecell{plate\\-slide-side} 
 & reach 
 & \makecell{reach\\-wall} 
 & \makecell{window\\-close} 
 & \makecell{window\\-open} 
 & \makecell{peg-unplug\\-side} 
 & -- & -- & -- & -- & -- & \\
 
\cmidrule(lr){2-12}
DP & \textbf{100.0 (0.0)} & 25.0 (5.0) & 53.3 (16.1) & \textbf{100.0 (0.0)} & 96.7 (5.8) & 15.0 (5.0) & -- & -- & -- & -- & -- & 79.9 \\
DP3 & \textbf{100.0 (0.0)} & 35.0 (13.2) & 61.7 (7.6) & \textbf{100.0 (0.0)} & \textbf{100.0 (0.0)} & \textbf{96.7 (2.9)} & -- & -- & -- & -- & -- & \textbf{92.3} \\
\rowcolor{gray!20}
\textbf{SCDP} & \textbf{100.0 (0.0)} & \textbf{41.7 (10.4)} & \textbf{80.0 (5.0)} & \textbf{100.0 (0.0)} & 98.3 (2.9) & 88.3 (2.9) & -- & -- & -- & -- & -- & 91.5 \\

\midrule
\multicolumn{13}{c}{\textbf{Meta-World Medium}} \\
\cmidrule(lr){1-13}
 & basketball 
 & \makecell{bin\\-picking} 
 & \makecell{box\\-close} 
 & \makecell{coffee\\-pull} 
 & \makecell{coffee\\-push} 
 & hammer 
 & \makecell{peg-insert\\-side} 
 & \makecell{push\\-wall} 
 & soccer 
 & sweep 
 & \makecell{sweep\\-into} 
 &  \\
 
\cmidrule(lr){2-12}
DP & 10.0 (0.0) & 23.3 (7.6) & 31.7 (7.6) & 58.3 (15.3) & 75.0 (5.0) & 35.0 (21.8) & 63.3 (17.6) & 31.7 (5.8) & \textbf{30.0 (10.0)} & 10.0 (0.0) & 28.3 (7.6) & 36.1 \\
DP3 & \textbf{100.0 (0.0)} & 53.3 (27.5) & 83.3 (2.9) & 93.3 (7.6) & 98.3 (2.9) & \textbf{100.0 (0.0)} & \textbf{88.3 (7.6)} & \textbf{98.3 (2.9)} & 28.3 (2.9) & \textbf{100.0 (0.0)} & 20.0 (17.3) & 78.5 \\
\rowcolor{gray!20}
\textbf{SCDP} & 71.7 (7.6) & \textbf{90.0 (0.0)} & \textbf{85.0 (0.0)} & \textbf{100.0 (0.0)} & \textbf{100.0 (0.0)} & \textbf{100.0 (0.0)} & 63.3 (10.4) & \textbf{98.3 (2.9)} & \textbf{30.0 (5.0)} & 93.3 (2.9) & \textbf{81.7 (5.8)} & \textbf{83.0} \\

\midrule
\multicolumn{13}{c}{\textbf{Meta-World Hard}} \\
\cmidrule(lr){1-13}
 & assembly 
 & \makecell{hand\\-insert} 
 & \makecell{pick-out\\-of-hole} 
 & \makecell{pick\\-place} 
 & push 
 & \makecell{push\\-back} 
 & -- & -- & -- & -- & -- &  \\
\cmidrule(lr){2-12}
DP & 15.0 (5.0) & 16.7 (7.6) & 11.7 (7.6) & 8.3 (5.8) & 20.0 (5.0) & 28.3 (12.6) & -- & -- & -- & -- & -- & 16.7 \\
DP3 & \textbf{100.0 (0.0)} & 28.3 (7.6) & 31.7 (7.6) & 85.0 (5.0) & \textbf{98.3 (2.9)} & 18.3 (2.9) & -- & -- & -- & -- & -- & 60.3 \\
\rowcolor{gray!20}
\textbf{SCDP} & 91.7 (2.9) & \textbf{81.7 (5.8)} & \textbf{80.0 (8.7)} & \textbf{86.7 (2.9)} & 93.3 (5.8) & \textbf{61.7 (12.6)} & -- & -- & -- & -- & -- & \textbf{82.5} \\

\midrule
\multicolumn{13}{c}{\textbf{Meta-World Very Hard}} \\
\cmidrule(lr){1-13}
 & \makecell{shelf\\-place} 
 & disassemble 
 & \makecell{stick\\-pull} 
 & \makecell{stick\\-push} 
 & \makecell{pick-place\\-wall} 
 & -- & -- & -- & -- & -- & -- &  \\
 
\cmidrule(lr){2-12}
DP & 25.0 (10.0) & 93.3 (2.9) & 33.3 (5.8) & 56.7 (7.6) & 21.7 (2.9) & -- & -- & -- & -- & -- & -- & 46.0 \\
DP3 & 60.0 (17.3) & \textbf{95.0 (8.7)} & 63.3 (10.4) & \textbf{100.0 (0.0)} & 95.0 (8.7) & -- & -- & -- & -- & -- & -- & 82.7 \\
\rowcolor{gray!20}
\textbf{SCDP} & \textbf{73.3 (7.6)} & \textbf{95.0 (0.0)} & \textbf{83.3 (2.9)} & \textbf{100.0 (0.0)} & \textbf{96.7 (5.8)} & -- & -- & -- & -- & -- & -- & \textbf{89.7} \\

\midrule
\multicolumn{13}{c}{\textbf{DexArt}} \\
\cmidrule(lr){1-13}
 & laptop 
 & faucet 
 & toilet 
 & bucket 
 & -- & -- & -- & -- & -- & -- & -- &  \\
\cmidrule(lr){2-12}
DP & 61.7 (2.9) & 41.1 (1.9) & 75.0 (1.7) & 49.4 (3.8) & -- & -- & -- & -- & -- & -- & -- & 56.8 \\
DP3 & 71.1 (5.9) & 38.9 (3.8) & 72.2 (2.5) & 28.3 (3.3) & -- & -- & -- & -- & -- & -- & -- & 52.6 \\
\rowcolor{gray!20}
\textbf{SCDP} & \textbf{77.8 (3.5)} & \textbf{45.0 (1.7)} & \textbf{75.6 (5.1)} & \textbf{51.1 (5.1)} & -- & -- & -- & -- & -- & -- & -- & \textbf{62.4} \\
\bottomrule
\end{tabular}
}
\end{table}

\subsection{Effect of a Wrist Camera}
\label{appendix:wrist_camera}
We evaluated SCDP with and without an additional wrist camera on six tasks.
Spatial conditioning can be applied to both
eye-to-hand and eye-in-hand cameras by projecting attention anchors into each view using the corresponding camera pose.
For the eye-in-hand setting, the projection matrix is updated as the end-effector moves.
As shown in Table~\ref{tab:wrist_camera}, using a wrist camera
slightly increased the average success rate from 82.5\% to 83.4\%.
Performance improved on Assembly, Hand-insert, and Push, but decreased on Pick-out-of-hole and Pick-place, indicating
task-dependent effects.

\begin{table}[h]
    \centering
    \caption{Success rates (\%) of SCDP with and without a wrist camera.}
    \label{tab:wrist_camera}
    \resizebox{\linewidth}{!}{%
        \begin{tabular}{lccccccc}
            \toprule
            & Assembly
            & Hand-insert
            & Pick-out-of-hole
            & Pick-place
            & Push
            & Push-back
            & Average \\
            \midrule
            
            SCDP (w/o Wrist)
            & 91.7
            & 81.7
            & \textbf{80.0}
            & \textbf{86.7}
            & 93.3
            & \textbf{61.7}
            & 82.5 \\

            \rowcolor{gray!20}
            SCDP (w/ Wrist)
            & \textbf{100.0}
            & \textbf{93.3}
            & 76.7
            & 71.7
            & \textbf{96.7}
            & \textbf{61.7}
            & \textbf{83.4} \\
            \bottomrule
        \end{tabular}%
        }
\end{table}

\clearpage

\begin{figure}[ht] 
    \centering
    \includegraphics[width=\linewidth]{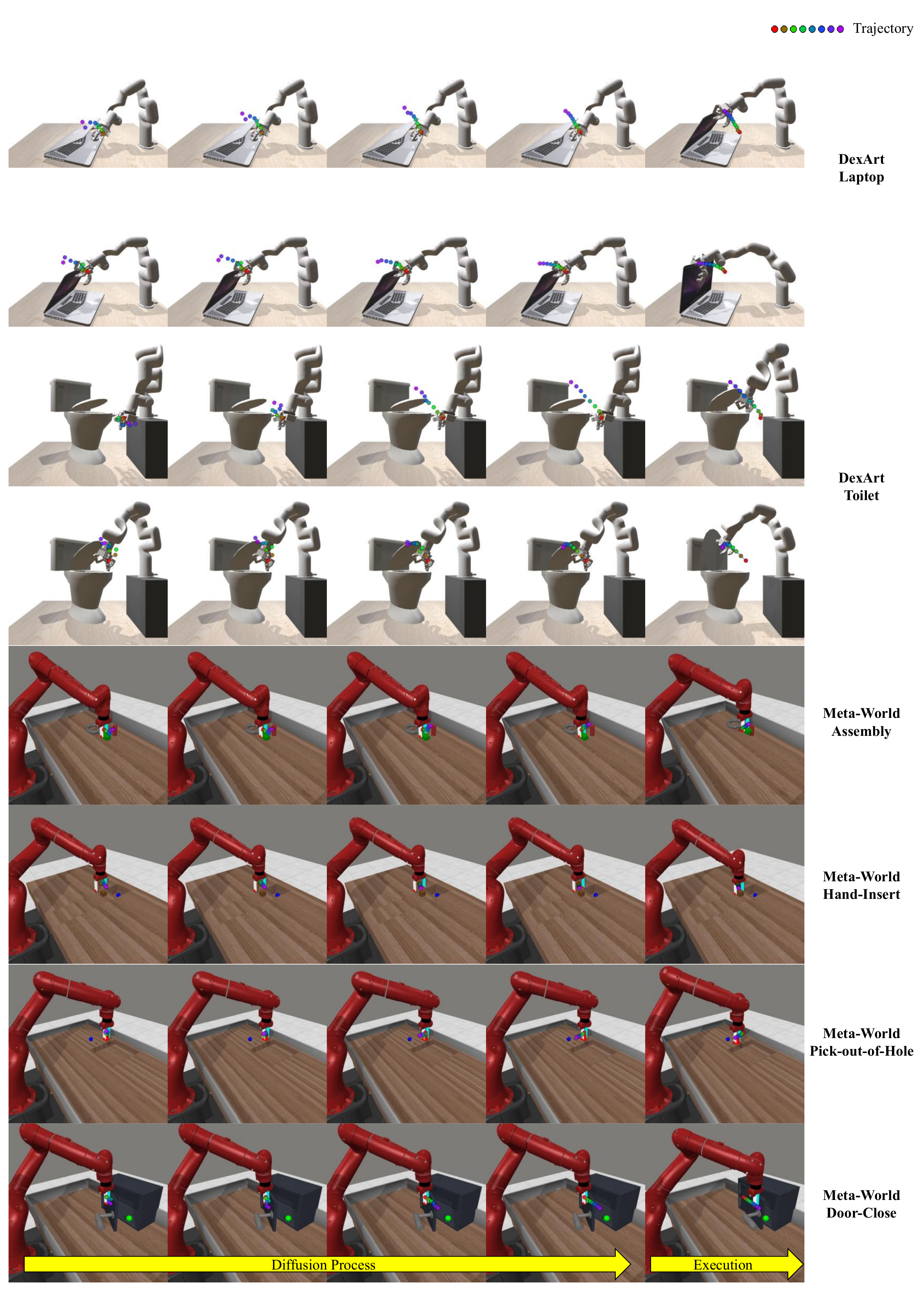}
    \caption{\textbf{Visualization of Spatial Conditioning.} SCDP extracts local visual features along the reconstructed trajectories, enabling the policy to focus on task-relevant interaction regions in broad scene images. The tool reference points are configurable across experimental settings; we use the palm point for DexArt and the gripper tool center point for Meta-World, as illustrated in Fig.~\ref{fig:SCDP_Framework}.}
    \label{fig:appendix_visualization}
\end{figure}

\clearpage

\section{Supplementary Materials: Real-World Experiments}

\subsection{Real-World Setup} 
\label{appendix:realworld_setup}
\begin{figure}[h] 
    \centering
    \includegraphics[width=\linewidth]{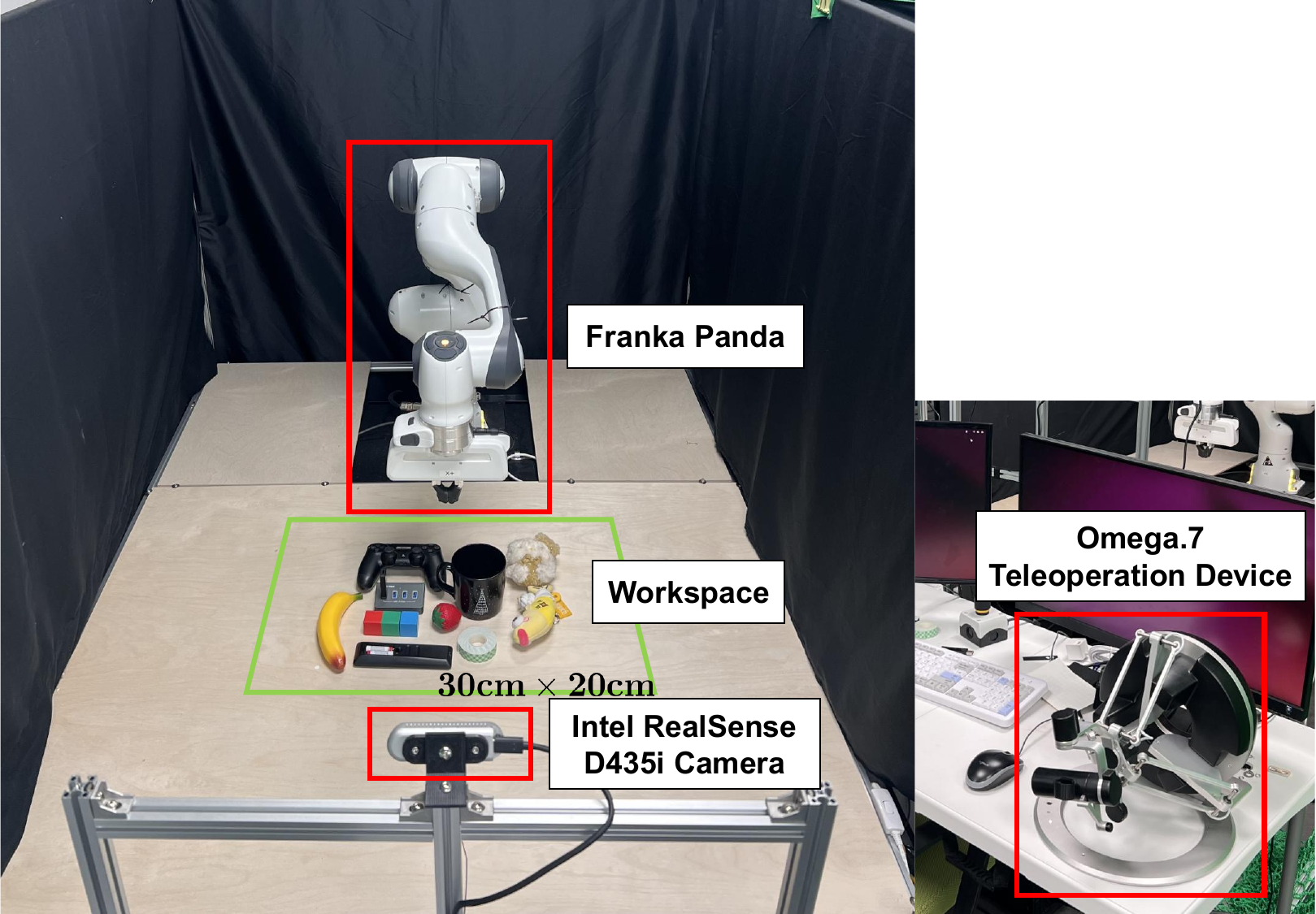}
    \caption{\textbf{Real-World Experimental Setup.}
    Franka Emika Panda robot with a parallel gripper, a single RGB camera, task objects, and a Force Dimension Omega.7 teleoperation device.}
    \label{fig:real_env}
\end{figure}

\begin{figure}[h] 
    \centering
    \includegraphics[width=1.0\linewidth]{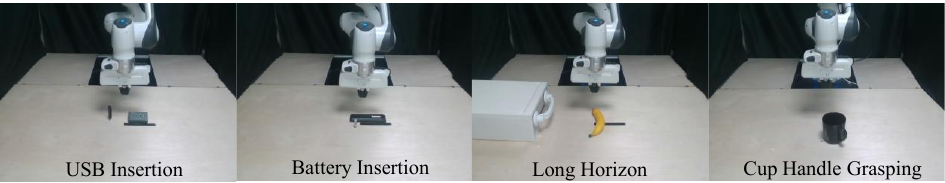}
    \caption{\textbf{Representative Camera Views.}
    Example images from collected demonstrations.}
    \label{fig:representative_views}
\end{figure}

We use a Franka Emika Panda robot equipped with a parallel gripper and a single Intel RealSense D435i camera for visual observations, as shown in Fig.~\ref{fig:real_env}. Expert demonstrations were collected via human teleoperation using a Force Dimension Omega.7 device. Representative camera views captured during demonstration collection are shown in Fig.~\ref{fig:representative_views}. The dataset comprises 180 demonstrations for \textit{USB Insertion}, 90 for \textit{Battery Insertion}, 120 for \textit{Long Horizon}, and 50 for each remaining task. For the \textit{USB Insertion} task, the hub port was placed at its initial position, and the USB drive pose was randomized within a radius of 5 cm. For the \textit{Battery Insertion} task, the battery slot was randomized along the black line shown in Fig.~\ref{fig:real_experiment}.  For all remaining tasks, the target objects and distractors were randomly placed within the $30\,\mathrm{cm} \times 20\,\mathrm{cm}$ workspace.

\clearpage

\subsection{Camera Calibration}
\begin{figure}[h] 
    \centering
    \includegraphics[width=0.5\linewidth]{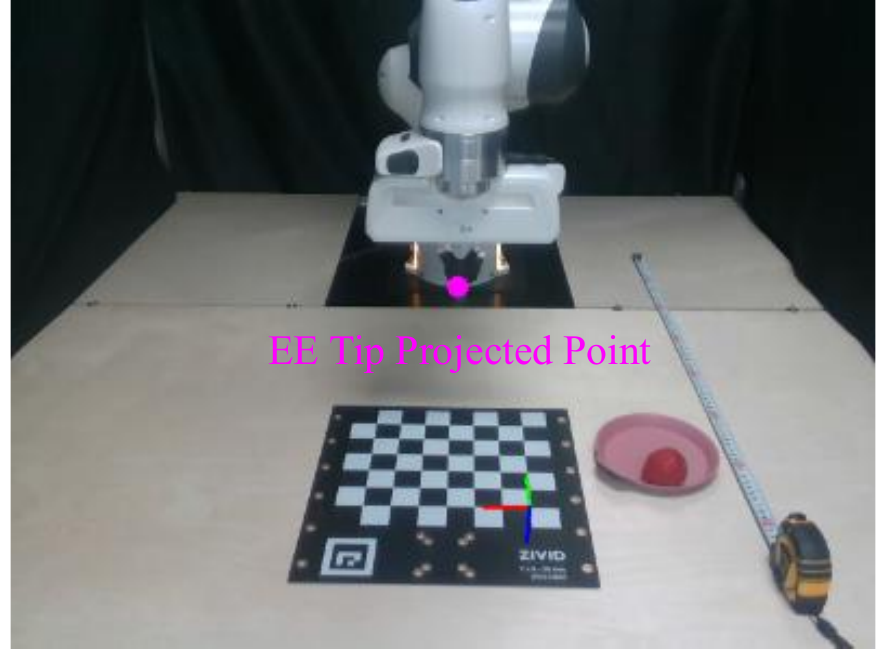}
    \caption{\textbf{Camera Calibration.}
    Projection of the end-effector tip onto the camera image, used to refine the camera-to-robot transformation.}
    \label{fig:projection}
\end{figure}

The spatial conditioning module uses the camera extrinsic parameters to project trajectory points expressed in the robot coordinate frame onto the image plane, as described in Section~\ref{subsec:spatial_conditioning}.
We obtained the \emph{camera-to-robot} transformation from the \emph{camera-to-board} and \emph{board-to-robot} transformations. 
The former was accurately estimated using a calibration board, while the latter was first approximated through manual measurement and then refined by projecting the end-effector tip into the image, as shown in Fig.~\ref{fig:projection}.
Calibration was performed only once and reused throughout all real-world experiments. 
Although slight projection shifts occasionally occurred due to factors such as robot vibration or measurement uncertainty, we did not observe noticeable performance degradation or sensitivity to these errors. 
These observations suggest some tolerance to small calibration errors in our setup, consistent with feature sampling along noisy intermediate action trajectories.

\subsection{Inference Efficiency}
At each denoising step, the intermediate trajectories are projected in a single batched GPU operation, and the corresponding visual features are bilinearly interpolated in parallel. We measured the inference latency of DP and SCDP using the same implementation-level optimizations. With 16 DDIM steps, generating a 16-step action prediction required 23.9 ms for DP and 27.9 ms for SCDP on an RTX 5090, corresponding to 4.0 ms of additional latency.

\end{document}